\documentclass[lettersize,journal]{IEEEtran}
\usepackage{amsmath,amsfonts}
\usepackage{algorithmic}
\usepackage{algorithm}
\usepackage{array}
\usepackage[caption=false,font=normalsize,labelfont=sf,textfont=sf]{subfig}
\usepackage{textcomp}
\usepackage{stfloats}
\usepackage{url}
\usepackage{verbatim}
\usepackage{graphicx}
\usepackage{cite}
\usepackage{multirow}

\newcommand{\eq}{Equation~}

\newcommand{\fig}{Fig.~}

\newcommand{\tab}{Table~}
\newcommand{\tabs}{Tables~}
\DeclareMathOperator{\conv}{Conv}
\DeclareMathOperator{\concat}{Concat}
\DeclareMathOperator{\mlp}{MLP}

\newcommand{\CP}[1]{\ignorespaces}
\newcommand{\ie}{\textit{i.e.~}}
\newcommand{\eg}{\textit{e.g.~}}

\begin{document}

\title{Multi-scale Feature Fusion with Point Pyramid for 3D Object Detection}

\author{Weihao Lu$^{1}$, Dezong Zhao$^{1}$,~\IEEEmembership{Senior Member,~IEEE}, Cristiano Premebida$^{2}$,~\IEEEmembership{Member,~IEEE}, Li Zhang$^{3}$,~\IEEEmembership{Senior Member,~IEEE}, Wenjing Zhao$^{4}$, Daxin Tian$^{5}$,~\IEEEmembership{Senior Member,~IEEE} 
        
\thanks{*This work was sponsored in part by the Engineering and Physical Sciences Research Council of the UK under the EPSRC Innovation Fellowship (EP/S001956/2) and in part by the Royal Society-Newton Advanced Fellowship (NAF\textbackslash R1\textbackslash 201213)   }
\thanks{$^{1}$Weihao Lu and Dezong Zhao are with the School of Engineering, University of Glasgow, University Avenue, Glasgow G12 8QQ, UK
        {\tt\small w.lu.1@research.gla.ac.uk, dezong.zhao@glasgow.ac.uk}}
\thanks{$^{2}$Cristiano Premebida is with the Institute of Systems and Robotics, University of Coimbra, 3030-290 Coimbra Portugal
        {\tt\small cpremebida@deec.uc.pt}}%
\thanks{$^{2}$Li Zhang is with the Department of Computer Science, Royal Holloway, University of London, Surrey, TW20 0EX, UK
        {\tt\small li.zhang@rhul.ac.uk}}%
\thanks{$^{4}$Wenjing Zhao is with the Department of Civil Environmental Engineering, Hong Kong Polytechnic University, Hong Kong, China
        {\tt\small wenjing.zhao@polyu.edu.hk}}
\thanks{$^{5}$Daxin Tian is with the School of Transportation Science and Engineering, Beihang University, Haidian District, Beijing 100191, China
        {\tt\small dtian@buaa.edu.cn}}
}



\maketitle

\begin{abstract}
Effective point cloud processing is crucial to LiDAR-based autonomous driving systems. The capability to understand features at multiple scales is required for object detection of intelligent vehicles, where road users may appear in different sizes. Recent methods focus on the design of the feature aggregation operators, which collect features at different scales from the encoder backbone and assign them to the \CP{regions or}points of interest. While efforts are made into the aggregation modules, the importance of how to fuse these multi-scale features has been overlooked. This leads to insufficient feature communication across scales. To address this issue, this paper proposes the Point Pyramid RCNN (POP-RCNN), a feature pyramid-based framework for 3D object detection on point clouds. POP-RCNN consists of a Point Pyramid Feature Enhancement (PPFE) module to establish connections across spatial scales and semantic depths for information exchange. The PPFE module effectively fuses multi-scale features for rich information without the increased complexity in feature aggregation. To remedy the impact of inconsistent point densities, a point density confidence module is deployed. This design integration enables the use of a lightweight feature aggregator, and the emphasis on both shallow and deep semantics, realising a detection framework for 3D object detection. With great adaptability, the proposed method can be applied to a variety of existing frameworks to increase feature richness, especially for long-distance detection. By adopting the PPFE in the voxel-based and point-voxel-based baselines, experimental results on KITTI and Waymo Open Dataset show that the proposed method achieves remarkable performance even with limited computational headroom.

\end{abstract}


\section{Introduction}
\IEEEPARstart{3}{D} object detection has attracted significant research interests as an important component of autonomous driving, especially when perception is the foundation of subsequent tasks in the autonomous driving framework \cite{9591277, 9917362}. 
Object detection in a two-dimensional space is commonly based on the textural information from RGB images, without knowing the distance to the objects. The 2D models can be extended with multi-view or depth estimation \cite{9689059} to realise detection in 3D space, without the usage of light-detection and ranging (LiDAR) sensors. However, no real depth measurement is involved. LiDAR sensors boost performance by providing accurate depth measurements. While the advantages are observed compared to detection on images, it is still challenging to process a point cloud due to its sparsity and irregularity. These limitations usually lead to sub-optimal detection results on faraway targets. In this paper, a Point Pyramid Region-Based Convolutional Neural Network (POP-RCNN) is proposed, which adopts the feature pyramid structure to encourage information exchange across spatial and semantic scales. The proposed Point Pyramid Feature Enhancement module effectively fuses multi-scale features to increase feature richness and ease the point distribution imbalance. While some research explores the multi-modality approach by fusing RGB images and point clouds \cite{f-pointnet,clocs,catdet,dvf,8370690, 9448387, 9511277}, it is still important to overcome the sparsity of pure point cloud representation. A richer point cloud-based feature representation can still improve the robustness of the system, especially in the case of failure of other sensors. Therefore, this paper focuses on pure point cloud input and presents the comparisons with the multi-modality methods on the Waymo and KITTI dataset. 

Recent research for 3D detection on point clouds builds on two main categories of frameworks, namely single-stage and two-stage. The single-stage detectors (SSDs) output detection results after a single encoding-decoding process \cite{3dssd,parta2,sassd,sessd,second,pointpillar}, while two-stage detectors (TSDs) perform extra refinement on the provisional results following the region proposal network (RPN) scheme \cite{pointrcnn,voxelrcnn,pvrcnn}. For both frameworks, feature extraction is usually realised by deploying the 3D encoder backbone, which consists of sparse and sub-manifold convolutions with different spatial strides. The increasing stride leads to a larger receptive field and encodes features at multiple scales. Most SSDs predict detection results on a 2D Bird's Eye View (BEV) feature map by flattening the 3D features of the last layer \cite{second}. Due to the finer-grained details exploited by the refinement stage, TSDs outperform SSDs in accuracy and gain more popularity even with heavier computation loads. A conventional refinement stage consists of a Region of Interest (RoI) pooling module together with the regression and classification heads. The RoI pooling module collects multi-scale features based on the proposal regions and further assigns the features to the anchor points inside each proposal. These features are stacked and fused by using fully connected layers before being fed to the regression and classification heads. However, the delicately encoded features are not used effectively. The combination of simple concatenation and fully connected layers is insufficient to establish communications across spatial scales and semantic depths, hence leading to a degraded capability of detecting distant objects. As a result, recent research tends to mystify other components of the model to improve detection accuracy on long range targets. However, this may lead to inefficient improvements by over-designing other components of the network. 


To remedy the scale imbalance, researchers explore the potential of Feature Pyramid Network (FPN) in the 2D detection domain \cite{fpn, panet, bifpn}. The uneven distribution of points in the 3D space further deteriorates this inefficiency in comparison with 2D detection. Thus a large variation in the number of points in objects remains a challenge. For example, in the KITTI dataset \cite{kitti}, the closest and furthest 10\% of objects have an average of around 350 and 15 points respectively. Voxel-FPN \cite{voxelfpn} is one of the early introductions of the pyramid structure to 3D point clouds. However, the integration of FPN with the voxel encoder backbone in a single-stage network results in a limited improvement over small targets.

To overcome the issues of ineffective feature fusion and point distribution imbalance, Point Pyramid RCNN (POP-RCNN) is proposed as a novel TSD network with a pyramid structure for 3D point clouds. POP-RCNN builds connections across various spatial scales and semantic depths for information exchange. In particular, the proposed method includes a 3D voxel backbone and a pyramid-based refinement network. The first stage 3D backbone uses sparse and sub-manifold convolutions to encode 3D features from voxelised volumes at multiple scales, followed by a 3D-to-BEV module which flattens the highly abstract 3D features to a 2D feature map. An intermediate 2D detection head is used to generate the regional proposals. The second stage is constructed with the Point Pyramid Pooling and Fusion (POP-Pool and POP-Fuse) modules. To aggregate features from multiple scales, the POP-Pool module sets up a collection of grid points for each proposal. Unlike the pooling modules in recent methods \cite{pvrcnn, pyramidrcnn}, POP-Pool aggregates the multi-scale features separately to each level of grid points. The POP-Fuse module follows the Generalised FPN \cite{giraffedet} to build spiderweb-like connections across all spatial scales and semantic depths, while adopting the point-specific three-nearest-neighbour (3NN) interpolation to match the feature size of different levels of POP-Pool. Combining POP-Pool with POP-Fuse, the PPFE module can fully exploit the encoded features by encouraging information exchange. The fused features carry dense semantic and geometry evidence to mitigate the imbalance of point distribution. To account for distance-invariant features, a distance-aware density confidence scoring (DADCS) scheme is integrated to provide additional guidance for classification confidence. POP-RCNN can be adopted in various TSDs. For example, voxel-based detector, Voxel-RCNN \cite{voxelrcnn}, and the point-voxel-based detector, PV-RCNN \cite{pvrcnn} are extended with the proposed approaches. Our POP-RCNN improves Voxel-RCNN and PV-RCNN by 2.88\% and 1.12\% on Vehicle LEVEL\_2 on Waymo Open Dataset respectively, and 0.63\% and 0.66\% on the Moderate Car category on KITTI \textit{val} set respectively. With the emphasis on long range detection, the proposed POP-RCNN improves its baseline by 2.02\%, 3.32\% and 1.02\% in mAP on LEVEL\_1 long distance ($>50$m) Vehicle, Pedestrian and Cyclist categories on Waymo Open Dataset respectively.

To summarise, the main contributions of this work are:
\begin{itemize}
    \item The POP-Pool and POP-Fuse modules are designed for point cloud processing by considering effective multi-scale feature representations. The separated pooling and fusion scheme can mitigate the point sparsity and distribution imbalance by ensuring sufficient information exchange across spatial and semantic dimensions.
    
    \item A new feature pyramid detection framework, POP-RCNN, is proposed for 3D object detection on point clouds. POP-RCNN integrates the POP-Pool and POP-Fuse to generate high-quality predictions on road users of different sizes, with the aid of DADCS scheme.
    
    \item Extensive experiments are presented, which prove the compatibility and effectiveness of POP-RCNN in improving existing two-stage 3D point cloud detectors.
\end{itemize}


\section{Related Works}

\subsection{3D Object Detection from Point Clouds}
Like object detection with 2D data representations, most existing 3D detectors can be categorised into SSDs and TSDs. SSDs propose a one-time bounding box generation based on the features encoded by the backbone network in detection models. As an example, VoxelNet \cite{voxelnet} utilises a voxel-based 3D backbone, where 3D CNNs can be applied to the 3D representations of regularised grids. SECOND \cite{second} improves the efficiency of 3D CNNs over sparse point clouds with sub-manifold and sparse convolutions. CIA-SSD \cite{ciassd} combines spatial and semantic features with attention fusion after the backbone network. SA-SSD \cite{sassd} builds an auxiliary network in parallel with the SECOND-based main branch. The auxiliary network inverts the voxelisation and calibrates the weights of the main branch with foreground segmentation and bounding box centre estimation. PointPillar \cite{pointpillar} reduces computation intensity by compressing the 3D scene into a pseudo-image in the $z$-direction. To avoid the manual design of anchor boxes, an anchor-free detection head is used by \cite{anchorfreessd} together with an Intersection-over-Union (IoU) calibration module. 3D-SSD \cite{3dssd} uses a point-based backbone originated from \cite{pointnet} to reduce partition effects induced by voxelisation. SE-SSD \cite{sessd} improves the detection accuracy of SSDs with self-ensemble networks. Despite the low inference time, the performance of SSDs is limited due to the lack of specific refinement to each bounding box. In comparison with the above existing studies, the proposed POP-RCNN in this research adopts the SECOND-based SSD backbone as the first stage RPN of the framework.

TSDs generate bounding boxes twice by extending the SSDs with an additional refinement module. The refinement module collects features with the guidance from the proposal generated by the first stage RPN, and further improves the detection with more fine-grained features. F-PointNet \cite{f-pointnet} collects subsets of points based on the proposal generated by a 2D detector from RGB images. The subsets of points are processed by a PointNet \cite{pointnet} based encoder for 3D bounding box refinement. Point-RCNN \cite{pointrcnn} generates proposals with a point-based backbone and performs region pooling for canonical 3D bounding box refinement. Voxel-RCNN \cite{voxelrcnn} integrates the voxel-based counterpart of region pooling to the SECOND backbone. PV-RCNN \cite{pvrcnn} assigns voxel features to key point coordinates to reduce the spatial errors induced by voxelisation. PDV \cite{PDV} incorporates voxel centroid localisation and point density awareness to a TSD framework. Semantic point generation (SPG) \cite{spg} recovers the lost foreground points by adopting an SPG module. SFD \cite{sfd} enhances raw point clouds with the extra depth information estimated from RGB images. However, the multi-scale features are not fully exploited. The refinement modules of these methods combine the aggregated features with concatenation and multi-layer perception (MLP). In contrast, the proposed POP-RCNN model in this research explores the potential of comprehensive multi-scale feature fusion, where feature density and richness are improved towards higher-quality bounding box generations.

\begin{figure*}[t]
\centering
\includegraphics[width=\textwidth]{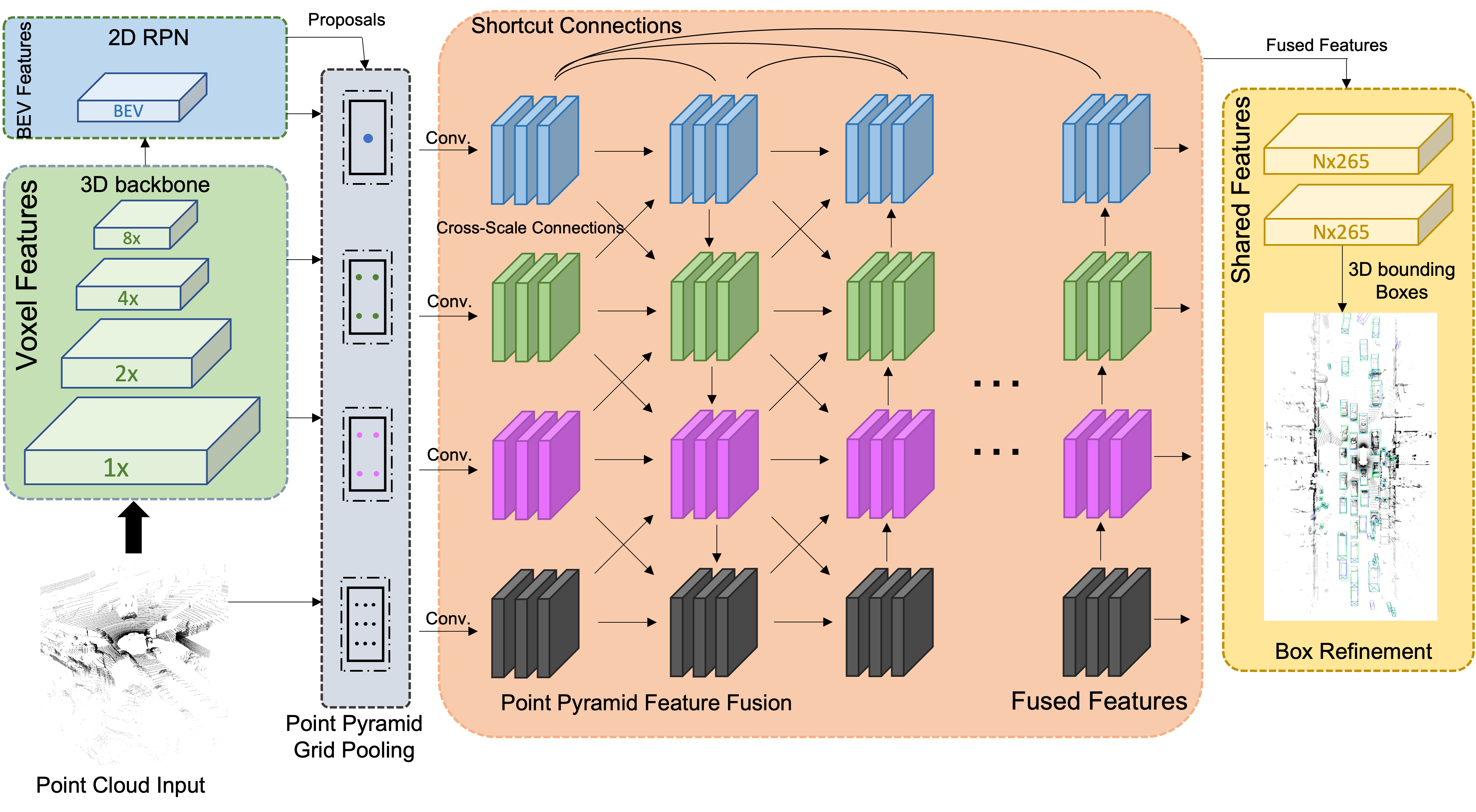}
\caption{Overview of the POP-RCNN architecture. In the first stage, the input point cloud is first voxelized and encoded with the 3D sparse convolution backbone to output $1\times$, $2\times$, $4\times$, $8\times$ voxel features. The 3D voxel features are flattened to $L\times W\times (H\times C)$, followed by a 2D RPN. The second stage of the network consists of POP-Pool, POP-Fuse and the detection head for box refinement. POP-Pool aggregates features with the guidance from proposal bounding boxes. Feature sources include raw points, voxel features ($1\times$, $2\times$, $4\times$ and $8\times$) and BEV features. POP-Fuse establishes a generalised FPN to fuse multi-scale features with cross-scale connections and shortcut connections. In the box refinement module, the fused features are processed with MLPs to compute the shared features for the regression and classification heads. With distance-aware density confidence scoring (DADCS), the final output of prediction bounding boxes is generated.}
\label{fig:archi}
\end{figure*}

\subsection{Feature Pyramid Network in Object Detection}
Feature Pyramid Structures are more popular in 2D image processing through multiple downsampling stages or strided convolutions. FPN \cite{fpn} adds a top-down path to the pyramid feature hierarchy to enrich the feature from shallow layers at higher resolutions. PANet \cite{panet} predicts all objects on a single feature map, which is generated by fusing feature maps at multiple scales with adaptive feature pooling and RoIAlign. ASFF \cite{asff} performs pyramidal feature fusion in conjunction with adaptive spatial feature fusion. FPT \cite{fpt} considers the non-local context of objects by using transformers for feature fusion. 
The introduction of a feature pyramid structure to 3D detectors is not trivial. Voxel-FPN \cite{voxelfpn} adopts a pyramidal structure in a SSD for fusing multi-scale voxel features. Pyramid-RCNN \cite{pyramidrcnn} deploys a pyramid RoI head as the refinement stage in a TSD. While Voxel-FPN \cite{voxelfpn} is limited in detecting small objects, Pyramid-RCNN \cite{pyramidrcnn} suffers from long inference time due to the multi-stage RoI-Gird Pooling and RoI-Grid Attention. In this research, the introduction of the Pyramid Feature Enhancement module improves the accuracy from its baseline models. Compared to existing methods, the proposed method consumes less memory with the simplified backbone.

\section{Methodology}
In this section, the POP-RCNN in introduced, which is a family of TSDs incorporating the point pyramid feature enhancement (PPFE). A typical POP-RCNN detector consists of a 3D backbone, a 2D region proposal network, an RoI pooling module and a detection head for generating predictions. By improving the RoI pooling module with the PPFE component, extra connections are built between feature nodes to provide sufficient communications across different scales and depths. In addition, the DADCS scheme is adopted for the detection head to consider the imbalanced point distribution. \fig\ref{fig:archi} illustrates the overall structure of POP-RCNN.

\subsection{3D Backbone Network}
To extract sufficient information from raw point inputs, we follow the configurations of recent methods \cite{voxelrcnn, pvrcnn}, where the 3D voxel backbone network \cite{second} is used. The input to a 3D detector is a set of 3D points, $\mathbf{P} = \{\mathbf{p}_i, \mathbf{f}_i \},\, i\in [1,\,N]$, where $\mathbf{p}_i$ and $\mathbf{f}_i$ are the coordinates and features of each individual point and $N$ is the total number of points. The point set $\mathbf{P}$ is first voxelised into a regularised representation. The feature values are calculated from $\mathbf{p}_i$ and $\mathbf{f}_i$ of the points in each voxel, while the feature location is given by the centre of that voxel. By applying the 3D sparse and sub-manifold convolutions, the 4 scales of 3D voxel features are obtained, which are with $1\times$, $2\times$, $4\times$ and $8\times$ downsampled resolutions compared to the original voxel map size respectively. The last layer's 3D features are flattened along the $z$-direction to produce a 2D BEV feature map. Similar to the image-based counterpart, a series of 2D convolution layers are formulated to encode abstract 2D features before feeding to the 2D detection head for RoIs and classification generation. These RoIs and classification scores are used as proposals for the subsequent refinement stage. For a point-voxel-based network \cite{pvrcnn}, point features are required instead of voxel features. Voxel features correspond to fictional voxel centres, while point features are assigned to actual point locations in the 3D space. In \cite{pvrcnn} the authors also compose an additional Voxel Set Abstraction module to associate the nearest voxel features to each sampled key point. To summarise, the outputs of the 3D backbone and the 2D RPN are: 1) 3D features, $\mathbf{F_{3D}} = [\mathbf{F}_{1\times}, \mathbf{F}_{2\times}, \mathbf{F}_{4\times}, \mathbf{F}_{8\times}]$, where $\mathbf{F_{3D}}$ is the voxel or point features for a voxel-based or point-voxel-based network; 2) 2D BEV features, $\mathbf{F_{BEV}}$, from the 2D RPN; 3) RoIs, $\mathbf{B}_{p} = [\mathbf{x}, \mathbf{y}, \mathbf{z}, \mathbf{l}, \mathbf{w}, \mathbf{h}, \mathbf{\Theta}]$, where $\mathbf{B}_{p}$ is a list of the proposed bounding boxes represented by the 3D locations of the box centres, dimensions and rotation angles of the boxes; 4) classification scores, $\mathbf{C}$, for the proposal bounding boxes, and 5) binary per-point masks, $\mathbf{M}$, to filter foreground and background points only in point-voxel-based networks.

\subsection{Point Pyramid Feature Enhancement}
In the refinement stage, the RoI pooling module collects multi-scale features from the 3D and BEV feature maps with guidance from the RoIs provided by the 2D RPN. In most existing methods \cite{pvrcnn,sasa,votr}, feature collection is achieved by assigning RoI grid points to each proposal. The RoI grid points are evenly distributed inside the proposal bounding boxes and serve as the downsampled key points, aggregating the features from neighbouring points or voxel centres. For a given set of grid points $\mathcal{G}\in \mathbb{R}^{3\times N_g}$, the pooled features can be expressed as: 

\begin{equation}
\label{eq:1}
    \mathbf{F'}_{\mathrm{pool}} = \mathcal{S}\, ([\mathbf{F}_{1\times}, \mathbf{F}_{2\times}, \mathbf{F}_{4\times}, \mathbf{F}_{8\times}, \mathbf{F_{BEV}}], \,  \mathbf{P_{key}}, \, \mathcal{G})
\end{equation}
where $\mathcal{S(\,\cdot\,)}$ is the aggregation function based on the maximum response or interpolation of the neighbouring features. $\mathbf{P_{key}}$ is the list of locations of the source features, which are the voxel centres in voxel-based networks or sampled key points in point-voxel-based networks. A single set of RoI grid points can provide sufficiently dense information in some scenarios. However, this limits the feature richness when objects are from distance, due to the non-uniform distribution and sparsity of point clouds.

To combine the pooled features, $\mathbf{F'}_{\mathrm{pool}}$, a simple concatenation is used, followed by a convolution layer with shared weights, 

\begin{equation}
    \mathbf{F}_{\mathrm{pool}} = \conv \, (\concat(\mathbf{F'}_{\mathrm{pool}}))
\end{equation}
where $\conv (\cdot)$ and $\concat (\cdot)$ are the convolution and concatenation operations respectively.

Despite the sophisticated design of a powerful backbone and aggregation module, the encoded features are not fully exploited. This leads to insufficient information exchange between features at different scales. Therefore, we introduce the pooling and fusion scheme inside PPFE.

\subsubsection{Point Pyramid Grid Pooling}
To mitigate the unfavourable effects of point clouds, a pyramid structure for point-based feature pooling is constructed, namely POP-Pool. Inspired by Pyramid-RCNN \cite{pyramidrcnn}, we sample multi-level grid points uniformly over each proposal box to ensure dense context information is captured at each spatial scale. The grid points of the $l$th level can be defined as:

\begin{equation}
    \mathcal{G}^l = \{ (x_i, y_j, z_k) \;|\; i \in[1,n^l_{x}], j \in [1,n^l_{y}], z\in [1,n^l_{z}] \}
\end{equation}
where $n^l_{x}$, $n^l_{y}$ and $n^l_{z}$ are the number of points in each dimension. Every level of the grid point pyramid in \cite{pyramidrcnn} is tasked to collect features from the same sources, which include the features at all scales. Instead of collecting features from all 3D feature voxel sets, POP-Pool separates the different feature sources and distributes the workload to each individual level of the pyramid. One level is only responsible for collecting features from one source, \eg features at one scale or the BEV features. This ensures that features from the same source are not pooled repeatedly. In contrast, in the Pyramid-RCNN \cite{pyramidrcnn}, all grid points are used to gather features from the combined feature map, which is generated via concatenating the encoded features from multiple scales. This leads to the duplication of feature pooling, hence more time and memory will be consumed. This indicates that the features from different spatial scales are actually combined via concatenation regardless of the pyramid structure. Furthermore, the subsequent pooling process in Pyramid-RCNN is not designated for multi-scale fusion. To this end, the integration of POP-Pool and POP-Fuse can ensure effectiveness in the subsequent fusion task by treating individual feature sources independently. Therefore, the number of sets of grid points in POP-Pool is designed to match the number of feature sources and the spacing between points is defined in accordance with the intermediate voxel sizes. Considering the different resolutions of the 3D voxels, the separate feature pooling scheme can improve efficiency by matching the grid sizes at each spatial scale level, as well as reducing redundant operations. More specifically, features at a lower resolution are aggregated by the level of sparser grid points (\ie smaller $(n^l_{x},n^l_{y},n^l_{z})$), which have a larger receptive field. \eq(\ref{eq:1}) can be transformed to represent the feature aggregation for the $l$th level of the grid point pyramid, as indicated in \eq(\ref{eq:4}):

\begin{equation}
\label{eq:4}
    \mathbf{F'}^l_{\mathrm{pool}} = \mathcal{S}\, (\mathbf{F}^l, \,  \mathbf{P}^l_\mathbf{key}, \, \mathcal{G}^l), \forall \; l\in [1, \,L]
\end{equation}
where $\mathbf{F}^l$ is one of the feature sources from $\mathbf{F_{3D}}$ and $\mathbf{F_{BEV}}$, and $L$ is the number of levels in the pyramid. The separated pooled features provide both rich contexts within and around the proposed bounding boxes, while maintaining a reasonable computational overhead. This design ensures feature richness by encouraging information exchange across multiple scales from a sparse data representation, where background points greatly outnumber the meaningful foreground points.

\begin{figure}[t]
\centering
\includegraphics[width=0.8\linewidth]{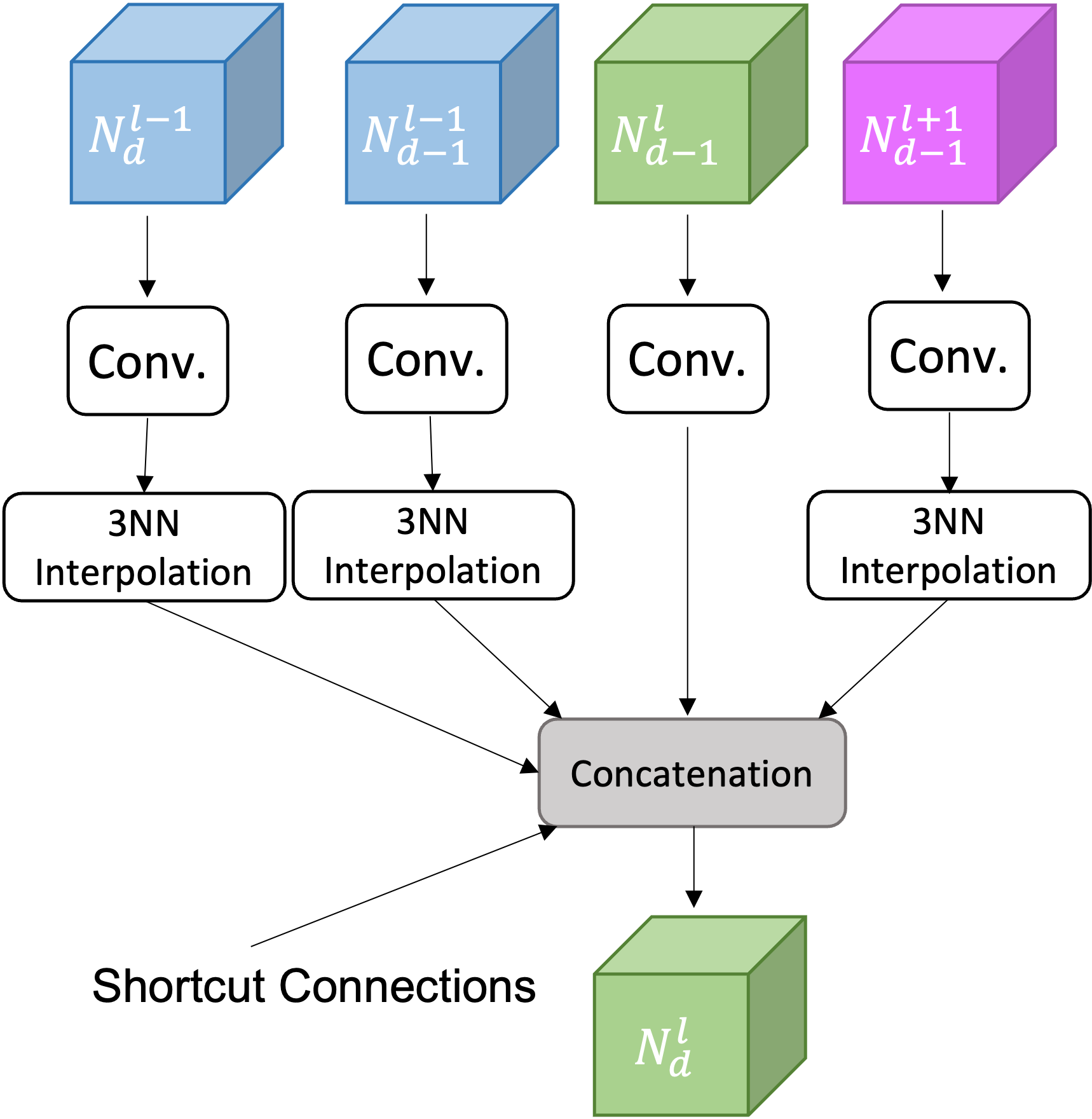}
\caption{Illustration of the detailed fusion scheme in POP-Fuse. For each feature node ($N^l_d$), the input nodes are determined based on the FPN structure depicted in \fig\ref{fig:archi} with cross-scale connections and shortcut connections. The features from the previous depth layer ($d-1$) are processed with convolutions. The features from the neighbouring level ($l\pm 1$) are resampled with 3NN interpolation as required to match the feature resolution. In addition, the shortcut connections are fed to the current node. All input features are combined with concatenation.}
\label{fig:fpns}
\end{figure}

\subsubsection{Point Pyramid Feature Fusion}
To fuse multi-scale features from POP-Pool, we deploy an FPN-based sub-network, namely POP-Fuse. VoxelFPN \cite{voxelfpn} constructs both top-down and bottom-up paths for fusing multi-level features from the voxel backbone, which lacks the cross-scale and shortcut connections. More cross-scale connections can encourage information exchange between different resolutions, while shortcut connections allow a large depth by mitigating the gradient degradation. Inspired by \cite{giraffedet}, we adopt the Generalised-FPN structure to connect a series of feature nodes as shown in \fig\ref{fig:fpns}. In particular, the pooled features are denoted as the first layer of feature nodes, $\mathbf{f}^l_0$. The shortcut connections are built to connect nodes on the same level of spatial scale. By using Dense Connections \cite{densely,semantic-mining} or $log_2n$ connections, the features from previous layers can be proceeded as:
\begin{equation}
    \mathit{f}^l_d = \conv (\concat (\mathit{f}^l_{\hat{d}})), \; 0\leq\hat{d}<d
\end{equation}
where the choice of $\hat{d}$ depends on the connection mode.

The cross-scale connections are built to connect nodes from the previous layer of adjacent scale levels. Due to the resolution difference between neighbouring levels, resampling the features to match the destination resolution is essential. Bilinear interpolation and max-pooling are popular in two-dimensional data. However, point-wise max-pooling leads to a loss of information, which becomes more severe in sparse 3D point clouds. To tackle this problem, we apply the 3NN interpolation for both upsampling and downsampling. The features enhanced by POP-Fuse can be summarised as:
\begin{equation}
    \mathbf{F}_{\mathrm{fused}} = \concat ([ \mathit{f}^1_D, \;..., \; \mathit{f}^L_D] )
\end{equation}
where $D$ is the number of total layers in depth. The fused features are fed to the bounding box regression and classification heads. 

\begin{figure}[t]
\centering
\includegraphics[width=\linewidth]{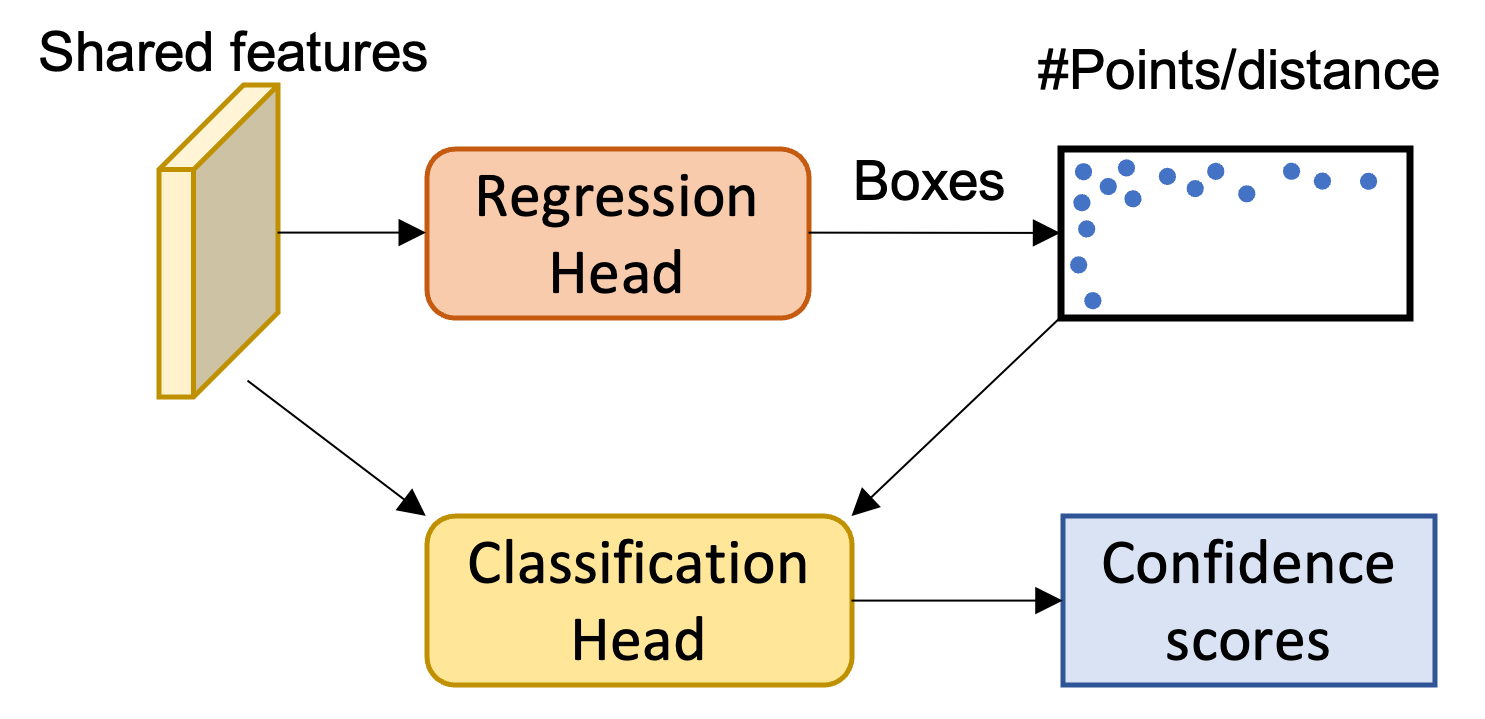}
\caption{Illustration of Distance-Aware Density Confidence Scoring. Bounding boxes are predicted with MLPs. According to the predictions, the number of points in each bounding box is summarised and the distance to the sensor is calculated. This information is combined with the shared features according to \eq (\ref{eq:dcs1}) and (\ref{eq:dcs2}). The classification head produces the calibrated confidence scores with MLPs.}
\label{fig:conf}
\end{figure}

With the aid of both cross-scale and shortcut connections, the fusion of multi-scale features is more sophisticated and can be extended in depth to explore high-dimensional semantics. The efficient combination of rich spatial information and high-dimensional semantics is essential to manage the feature sparsity and large scale variations in the dataset.

\subsection{Distance-Aware Density Confidence Scoring}
To further solve the sparsity and non-uniform distribution of points, a DADCS scheme is deployed. In addition to the fused features, the information about the point distribution inside the bounding box is exploited. A strategy that includes number of points in the feature map is adopted by \cite{PDV}. The proposed distance-aware density confidence scoring scheme further addresses the density variation in point clouds, by considering the relation between distance and point sparsity. Due to the nature of a LiDAR sensor, the point density decreases as the radius increases in cylindrical coordinates centred at the sensor. 
Distant objects tend to have a lower response in the feature map. However, it is important to maintain feature invariance with the attempt to accurately detect objects in a wide range of distances. This concept follows the scale invariance of 2D features in object detection on images. In particular, we compute the extra features from the number of points in the output bounding box, which is predicted by the regression head. With multi-layer perceptron denoted by $\mlp(\cdot)$, the regression head can proceed as follows:
\begin{equation}
    \mathbf{b} = \mlp_{\mathrm{reg}} (\mathbf{F}_{\mathrm{shared}})
\end{equation}
where $\mathbf{F}_{\mathrm{shared}} = \mlp (\mathbf{F}_{\mathrm{fused}})$. Given the bounding boxes $\mathbf{b}$, the extra density features can be computed by the Hadamard product of the logarithm of the numbers of points in the bounding boxes and the distances of the bounding boxes to the sensor, expressed as:
\begin{equation}
\label{eq:dcs1}
    \mathit{f}_{\mathrm{density}} = \log (\mathbf{N_b}) \circ \mathbf{s}
\end{equation}
Therefore the classification scores are calibrated with the additional features by:
\begin{equation}
\label{eq:dcs2}
    \mathbf{c} = \mlp_{\mathrm{cls}} ( \concat(   [\mathit{f}_{\mathrm{density}}  , \; \mathbf{F}_{\mathrm{shared}}]) )
\end{equation}

\subsection{Training Loss}
The POP-RCNN is trained end-to-end by the RPN loss $\mathcal{L}_{\mathrm{RPN}}$ and the refinement network loss $\mathcal{L}_{\mathrm{RCNN}}$. By denoting the predicted and ground truth classification results and bounding box residuals as ${c'_i}$, $\hat{{c_i}}$, $\delta b'_i$ and $\delta\hat{b_i}$ respectively, the loss function for the RPN can be expressed by:
\begin{multline}
\mathcal{L}_{\mathrm{RPN}} = \frac{1}{N_p} \left[ \sum_i \mathcal{L}_{\mathrm{cls}}({c'_i}, \hat{{c_i}}) + \right.\\
\left. \mathbb{I}(\mathrm{IoU}_i>\tau')\sum_i \mathcal{L}_{\mathrm{reg}} (\delta b'_i, \delta \hat{{b_i}})\right]
\end{multline}
where $N_p$ is the number of proposals. By considering the refined outputs ${c_i}$ and $\delta b_i$, the loss function for the refinement stage can be expressed by:
\begin{multline}
\mathcal{L}_{\mathrm{RCNN}} = \frac{1}{N_{\mathrm{box}}} \left[ \sum_i \mathcal{L}_{\mathrm{cls}}({c_i}, \hat{{c_i}}) + \right.\\
\left. \mathbb{I}(\mathrm{IoU}_i>\tau)\sum_i \mathcal{L}_{\mathrm{reg}} (\delta b_i, \delta \hat{{b_i}})\right]
\end{multline}
where $N_{\mathrm{box}}$ is the number of final bounding boxes. By setting the IoU threshold to $\tau'$ and $\tau$ for the two stages respectively, only the regression loss of the positive bounding boxes is included. $\mathcal{L}_{\mathrm{cls}}$ and $\mathcal{L}_{\mathrm{reg}}$ are cross-entropy loss and smooth-L1 loss respectively. The total loss is calculated by
\begin{equation}
    \mathcal{L} = \mathcal{L}_{\mathrm{RPN}} + \mathcal{L}_{\mathrm{RCNN}}.
\end{equation}

\section{Experiments}
In this section, the performance of POP-RCNN is evaluated on the Waymo Open Dataset \cite{waymo} and KITTI dataset \cite{kitti}. A brief introduction to datasets and experimental settings is given in Section \ref{sec:dataset}. We then compare POP-RCNN with other state-of-the-art methods on both datasets. Finally, the ablation studies are conducted to analyse the effectiveness of the components in the framework.

\subsection{Dataset}
\label{sec:dataset}
\textbf{Waymo Open Dataset}. WOD is a large dataset with 798 training sequences and 202 validation sequences, with around 160k and 40k point cloud samples respectively. Annotations are given to the 360{\textdegree} field of view. The annotations cover a range of $[-75.2, \, +75.2]$m, $[-75.2, \, +75.2]$m and $[-2, \, +4]$m in $X$, $Y$ and $Z$ axes respectively. The evaluation metrics are calculated as the mean Average Precision (mAP) and the mean Average Precision weighted by Heading (mAPH). The 3D intersection-over-union (IoU) thresholds for the bounding boxes are $(0.7, 0.5, 0.5)$ for Car, Pedestrian and Cyclist categories. Depending on how the testing samples are split, the results can be formatted by difficulty levels and detection ranges. By difficulty levels, ground truth targets are divided into LEVEL\_1 and LEVEL\_2, which guarantees at least 5 and 1 laser points are reflected from the objects. By detection ranges, the ground truth targets are assigned to the groups of $0-30\mathrm{m}$, $30-50\mathrm{m}$ and $>50\mathrm{m}$ from the sensor.

\textbf{KITTI dataset}. The KITTI 3D object detection dataset contains 7481 and 7518 samples for training and testing respectively. The training set is further divided into \textit{train} and \textit{val} splits with 3712 and 3769 samples respectively. Annotations are only given to the objects in the front, which cover a range of $[0, \, +70.4]$m, $[-40, \, +40]$m and $[-3, \, +1]$m in $X$, $Y$ and $Z$ axes respectively. The official evaluation metric is the mAP calculated by the official evaluation tool with 40 points from the precision-recall curve on three difficulty levels. The 3D IoU thresholds are $(0.7, 0.5)$ for Car and Cyclist categories. 

\begin{table*}[t!]
\centering
\caption{Performance comparison on the Waymo Open Dataset with 202 validation sequences by object types.}
\begin{tabular}{c|cccc|cccc|cccc}
\hline
\multirow{2}{*}{Methods} &
  \multicolumn{2}{c}{Veh. LEVEL\_1} &
  \multicolumn{2}{c|}{Veh. LEVEL\_2} &
  \multicolumn{2}{c}{Ped. LEVEL\_1} &
  \multicolumn{2}{c|}{Ped. LEVEL\_2} &
  \multicolumn{2}{c}{Cyc. LEVEL\_1} &
  \multicolumn{2}{c}{Cyc. LEVEL\_2} \\
            & mAP   & mAPH  & mAP   & mAPH  & mAP   & mAPH  & mAP   & mAPH  & mAP   & mAPH  & mAP   & mAPH  \\ \hline
SECOND \cite{second}      & 72.27 & 71.69 & 63.85 & 63.33 & 68.7  & 58.18 & 60.72 & 51.31 & 60.62 & 59.28 & 58.34 & 57.05 \\
PillarNet-34 \cite{shi2022pillarnet}      & 79.10  &  78.60  &  70.90  &  70.50  &  80.60  &  74.00  &  72.30  &  66.20  &  72.30  &  71.20  &  69.70  &  68.70 \\
Part-A2 \cite{parta2}     & 74.82 & 74.32 & 65.88 & 65.42 & 71.76 & 63.64 & 62.53 & 55.3  & 67.35 & 66.15 & 65.05 & 63.89 \\
PV-RCNN++ (1f) \cite{pvrcnn++}   & \textbf{79.25} &   \textbf{78.78} &  \textbf{70.61} &   \textbf{70.18}   &   \textbf{81.83} &  \textbf{76.28}&   73.17&   68.00    &   73.72 &  72.66 &  \textbf{71.21} &  \textbf{70.19}  \\
PDV \cite{PDV}         & 76.85 & 76.33 & 69.30 & 68.81 & 74.19 & 65.96 & 65.85 & 58.28 & 68.71 & 67.55 & 66.49 & 65.36 \\
VoxSet \cite{voxset}     & 77.82 & -     & 70.21 & -     & -     & -     & -     & -     & -     & -     & -     & -     \\
BtcDet \cite{btcdet}     & 78.58 & 78.06 & 70.10 & 69.61 & -     & -     & -     & -     & -     & -     & -     & -     \\
Pyramid-PV \cite{pyramidrcnn} & 76.3  & 75.68 & 67.23 & 66.68 & -     & -     & -     & -     & -     & -     & -     & -     \\ 
CasA \cite{casa}  & 78.55 & 78.06     & 69.67 & 69.23     & 77.22     & 70.60     & 68.06     & 62.08     & 68.19     & 66.76     & 65.73     & 64.33     \\
CenterPoint \cite{yin2021center} &   76.70 &     76.20  &  68.80 &  68.30 &  79.00 &  72.90 & 71.00  & 65.30 & -     & -     & -     & -  \\
SST\_TS (1f) \cite{fan2022embracing}  & 76.22 & 75.79     & 68.04 & 67.64     & 81.39     & 74.05     & 72.82     & 65.93     & -     & -     & -     & -     \\
CenterFormer (1f) \cite{zhou2022centerformer}  & 75.20 & 74.70     & 70.20 & 69.70     & 78.60     & 73.00     & 73.60     & 68.30     & 72.30     & 71.30     & 69.80     & 68.80     \\
M3DETR \cite{guan2022m3detr}  & 75.71 & 75.08     & 66.58 & 66.02     & -     & -     & -     & -     & -     & -     & -     & -     \\\hline

Voxel-RCNN \cite{voxelrcnn}  & 75.59 & -     & 66.59 & -     & -     & -     & -     & -     & -     & -     & -     & -     \\
\textbf{POP-RCNN-V}      & 77.89 & 77.42 & 69.47 & 69.04 & 80.04 & 74.06 & 71.58 & 66.01 & \textbf{73.75} & \textbf{72.68} & 71.04 & 69.98 \\ \hline
PV-RCNN† \cite{pvrcnn}     & 78.00 & 77.50 & 69.43 & 68.98 & 79.21 & 73.03 & 70.42 & 64.72 & 71.46 & 70.27 & 68.95 & 67.79 \\
\textbf{POP-RCNN-PV}        & 78.84 & 78.37 & 70.55 & 70.13 & 81.68 &  76.17 & \textbf{73.43} & \textbf{68.22} & 73.52 & 72.38 & 70.83 & 69.73 \\ \hline
\end{tabular}
\label{tab:wod_obj}
\end{table*}

\begin{table}[t]
\centering
\caption{Performance comparison on the Waymo Open Dataset with 202 validation sequences for vehicle class by range.  }
\begin{tabular}{c|ccc}
\hline
\multirow{2}{*}{Methods} & \multicolumn{3}{c}{Vehicle LEVEL\_1 mAP/mAPH} \\
                         & 0-30m         & 30-50m        & 50m-inf       \\ \hline
Part-A2 \cite{parta2}                  & 92.35/-       & 75.91/-       & 54.06/-       \\
CT3D \cite{ct3d}                     & 92.51/-       & 75.07/-       & 55.36/-       \\ 
PV-RCNN++ (1f) \cite{pvrcnn++}               & 93.34/-       & \textbf{78.08}/-       & 57.19/-       \\
Pyramid-PV \cite{pyramidrcnn}      & 92.67/92.20   & 74.91/74.21       & 54.54/53.45   \\
VoxSet \cite{voxset}               & 92.5/-        & 70.10/-           & 43.20/-       \\
PDV \cite{PDV}                     & 93.13/\textbf{92.71}   & 75.49/74.91   & 54.75/53.90   \\
BtcDet \cite{btcdet}              & \textbf{96.11}/-       & 77.64/-       & 54.45/-       \\
M3DETR \cite{guan2022m3detr}        &  92.69/92.22 &    73.65/72.94 & 52.96/51.80  \\
\hline
Voxel-RCNN \cite{voxelrcnn}              & 92.49/-       & 74.09/-       & 53.15/-       \\
\textbf{POP-RCNN-V}                    & 92.74/92.33   & 77.05/76.53   & 57.40/56.70  \\ \hline
PV-RCNN \cite{pvrcnn}                  & 92.96/-       & 76.47/-       & 55.96/-       \\
\textbf{POP-RCNN-PV}                    & 92.95/92.57   & 77.72/\textbf{77.21}   & \textbf{57.98}/\textbf{57.26}   \\ \hline \\ \hline
\multirow{2}{*}{Methods} & \multicolumn{3}{c}{Vehicle LEVEL\_2 mAP/mAPH} \\
                         & 0-30m         & 30-50m        & 50m-inf       \\ \hline
CT3D \cite{ct3d}                    & 91.76/-       & 68.93/-       & 42.60/-       \\
PDV \cite{PDV}                     & 92.41/\textbf{91.99}   & 69.36/68.81   & 42.16/41.48   \\ 
BtcDet \cite{btcdet}              & \textbf{95.99}/-       & 70.56/-       & 43.87/-      \\ 
M3DETR \cite{guan2022m3detr}        &  91.92/91.45 &    65.73/65.10  & 40.44/39.52  \\\hline
Voxel-RCNN \cite{voxelrcnn}              & 91.74/-       & 67.89/-       & 40.80/-       \\
\textbf{POP-RCNN-V}                     & 91.54/91.13   & 70.78/70.30   & 44.96/44.39  \\ \hline
\textbf{POP-RCNN-PV}                     & 91.79/91.41   & \textbf{71.42}/\textbf{70.94}   & \textbf{45.46}/\textbf{44.87}  \\ \hline
\end{tabular}
\label{tab:wod_veh_range}
\end{table}

\begin{table}[t]
\centering
\caption{Performance comparison on the Waymo Open Dataset with 202 validation sequences for pedestrian class by range. }
\begin{tabular}{c|ccc}
\hline
\multirow{2}{*}{Methods} & \multicolumn{3}{c}{Pedestrian LEVEL\_1 mAP/mAPH} \\
                         & 0-30m         & 30-50m        & 50m-inf       \\ \hline
Part-A2 \cite{parta2}                & 81.87/-       & 73.65/-       & 62.34/-       \\
PV-RCNN++ (1f) \cite{pvrcnn++}              & 84.88/-       & 79.65/-       & 70.64/-       \\
PDV \cite{PDV}                     & 80.32/73.60   & 72.97/63.28   & 61.69/50.07   \\ \hline
\textbf{POP-RCNN-V}                     & {85.30}/{80.83}   & {79.75}/{73.13}   & 71.77/62.44   \\ \hline
PV-RCNN \cite{pvrcnn}                & 83.33/-       & 78.53/-       & 69.36/-       \\
\textbf{POP-RCNN-PV}                     & \textbf{86.22}/\textbf{81.91}   & \textbf{80.50}/\textbf{74.09}   & \textbf{72.68}/\textbf{63.63}   \\ \hline \\ \hline
\multirow{2}{*}{Methods} & \multicolumn{3}{c}{Pedestrian LEVEL\_2 mAP/mAPH} \\
                         & 0-30m         & 30-50m        & 50m-inf       \\ \hline
PDV \cite{PDV}                     & 75.26/68.82   & 65.78/56.85   & 47.46/38.30   \\ \hline
\textbf{POP-RCNN-V}                     & {80.73}/{76.38}   & {72.56}/{66.37}   & {57.46}/{49.46}   \\
\textbf{POP-RCNN-PV}                     & \textbf{81.65}/\textbf{77.51}   & \textbf{73.41}/\textbf{67.38}   & \textbf{58.65}/\textbf{50.88}   \\\hline
\end{tabular}
\label{tab:wod_ped_range}
\end{table}

\begin{table}[t]
\centering
\caption{Performance comparison on the Waymo Open Dataset with 202 validation sequences for cyclist class by range.  }
\begin{tabular}{c|ccc}
\hline
\multirow{2}{*}{Methods} & \multicolumn{3}{c}{Cyclist LEVEL\_1 mAP/mAPH} \\
                         & 0-30m         & 30-50m        & 50m-inf       \\ \hline
Part-A2 \cite{parta2}                & 80.87/-       & 62.57/-       & 45.04/-       \\
PV-RCNN++ (1f) \cite{pvrcnn++}              & \textbf{83.65}/-       & 68.90/-       & 51.41/-       \\
PDV \cite{PDV}                     & 80.86/79.83   & 62.61/61.45   & 46.23/44.12   \\ \hline
\textbf{POP-RCNN-V}                    & 83.25/\textbf{82.23}   & \textbf{69.02}/\textbf{68.21}   & {53.46}/{52.11}   \\ \hline
PV-RCNN \cite{pvrcnn}                & 81.10/-       & 65.65/-       & 52.58/-       \\
\textbf{POP-RCNN-PV}                    & 82.94/{81.84}   & {68.44}/{67.20}   & \textbf{53.60}/\textbf{52.16}   \\ \hline \\ \hline
\multirow{2}{*}{Methods} & \multicolumn{3}{c}{Cyclist LEVEL\_2 mAP/mAPH} \\
                         & 0-30m         & 30-50m        & 50m-inf       \\ \hline
PDV \cite{PDV}                     & 80.42/79.40   & 58.95/57.87   & 43.05/41.09   \\ \hline
\textbf{POP-RCNN-V}                    & \textbf{82.66}/\textbf{81.64}   & \textbf{65.19}/\textbf{64.05}   & \textbf{49.81}/{48.55}   \\
\textbf{POP-RCNN-PV}                    & {82.36}/{81.26}   & {64.73}/{63.55}   & {49.55}/\textbf{48.61}   \\ \hline
\end{tabular}
\label{tab:wod_cyc_range}
\end{table}

\begin{figure*}[t]
\centering
\subfloat[]{\includegraphics[width=0.45\linewidth]{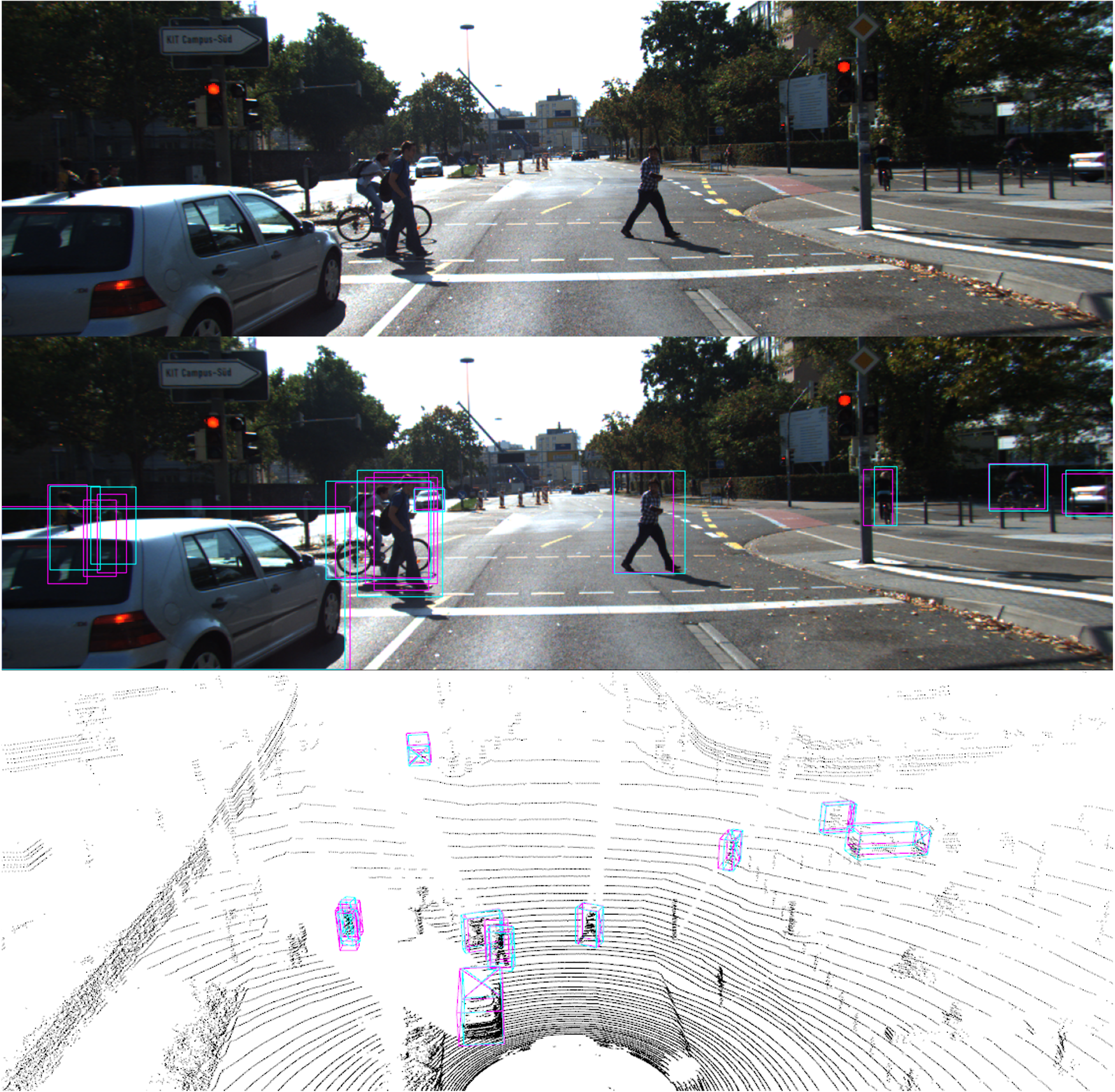}%
\label{fig:viz-a}}
\hfil
\subfloat[]{\includegraphics[width=0.45\linewidth]{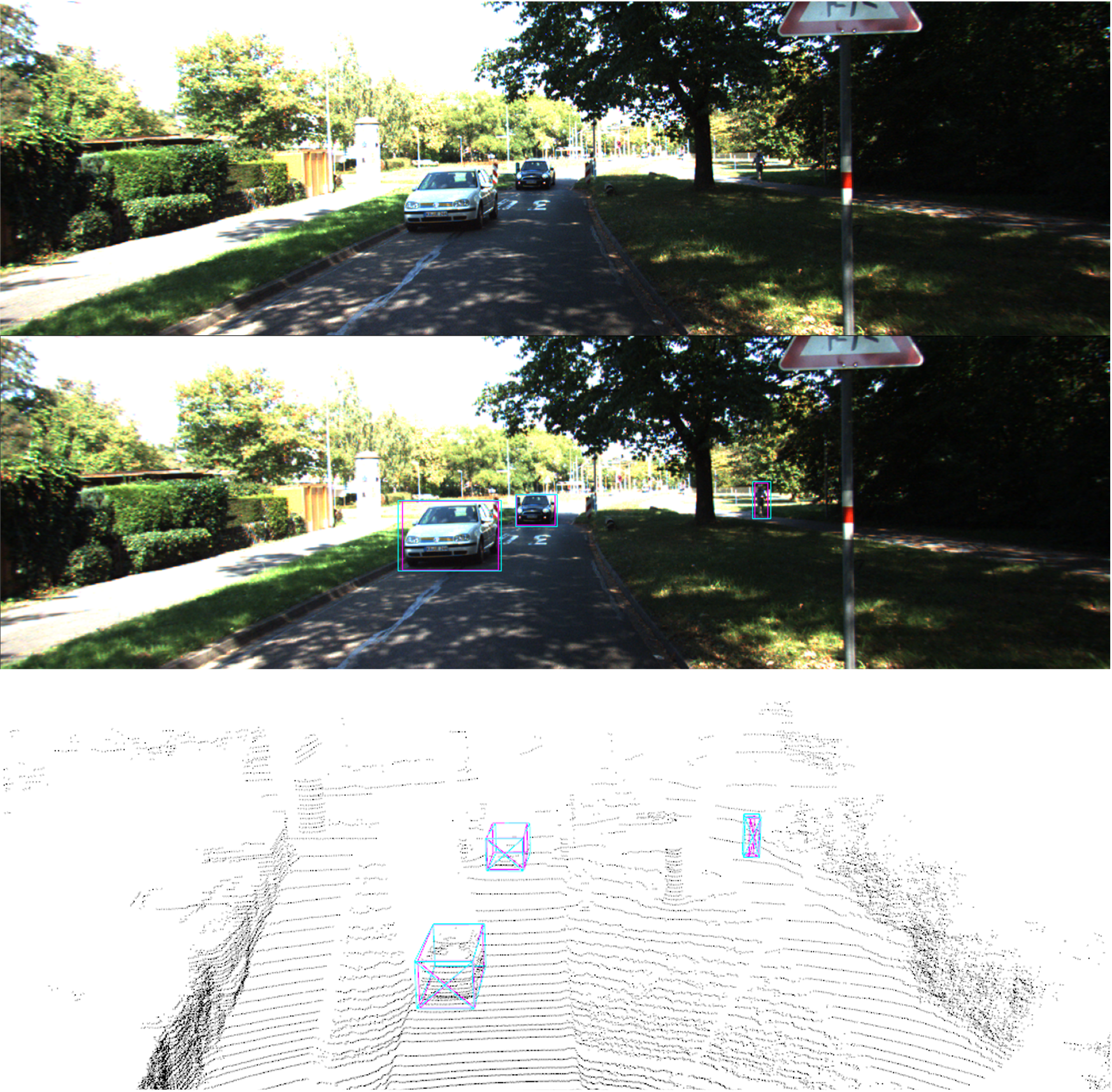}%
\label{fig:viz-b}}
\hfil
\subfloat[]{\includegraphics[width=0.45\linewidth]{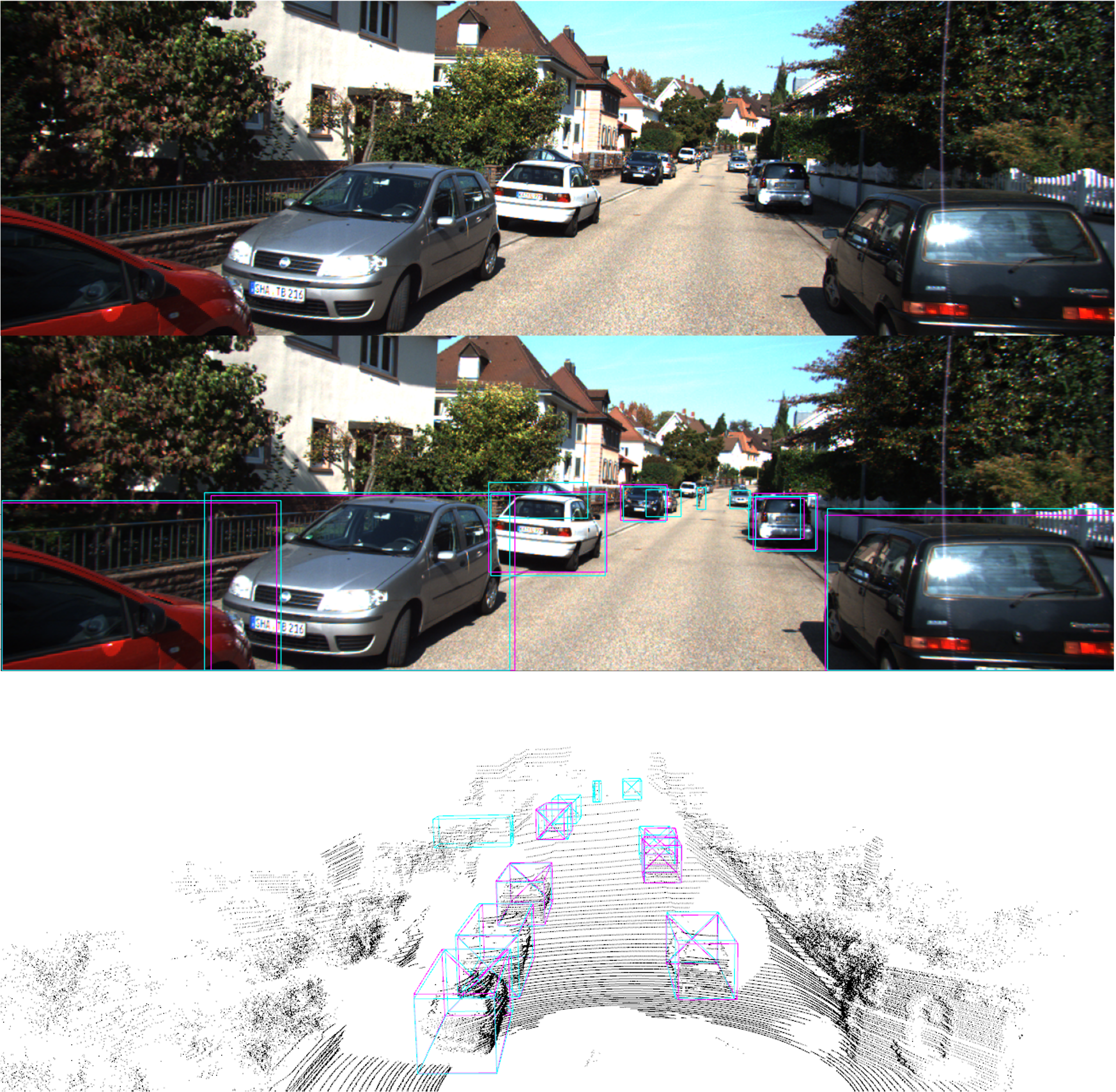}%
\label{fig:viz-c}}
\hfil
\subfloat[]{\includegraphics[width=0.45\linewidth]{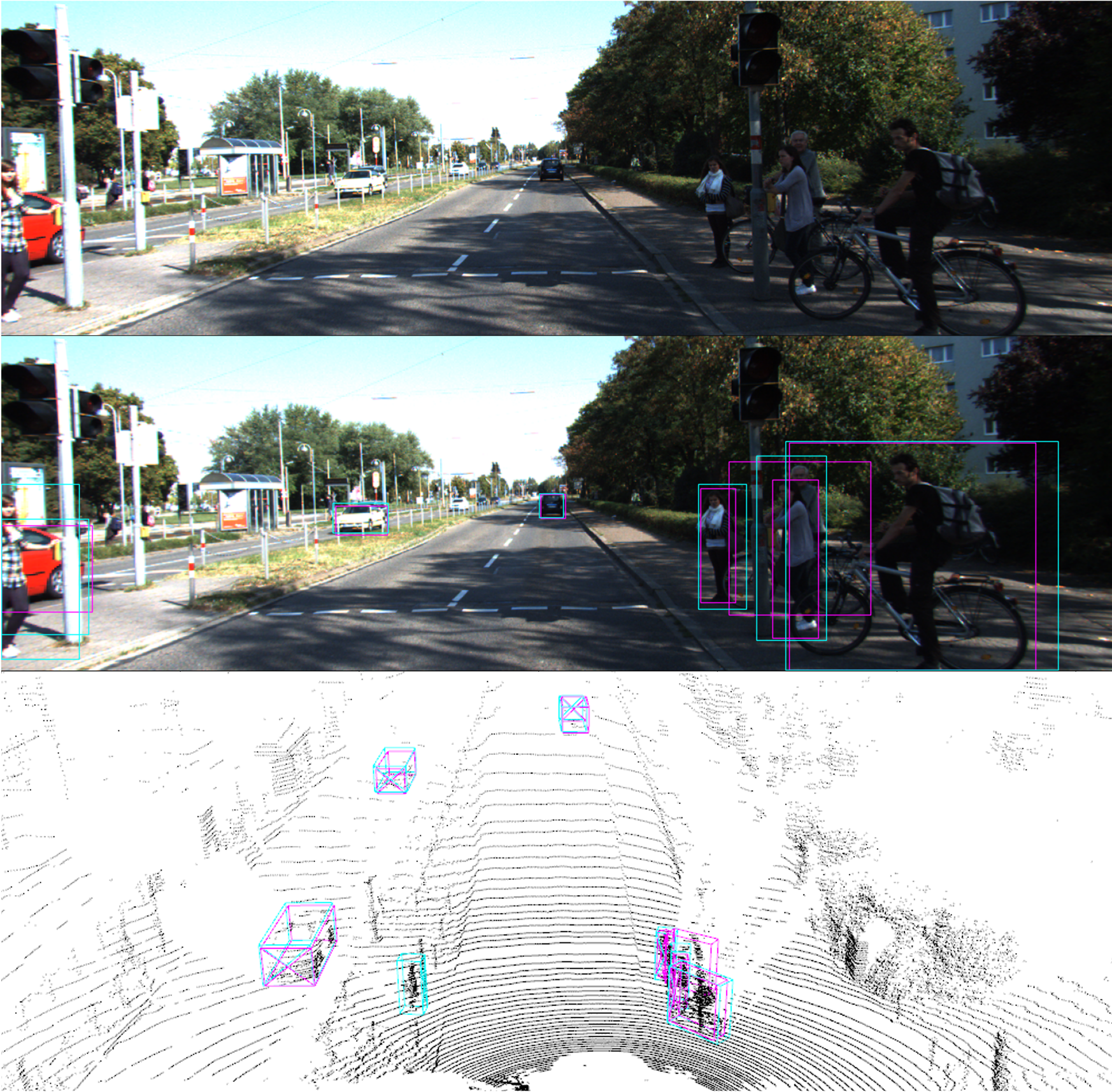}%
\label{fig:viz-d}}
\caption{Visualisations of results on KITTI \textit{val} split. The original images, images with annotations and point clouds with annotations are presented at the first, second and third rows respectively. Blue boxes indicate predictions and purple boxes indicate the ground truths. In \fig\ref{fig:viz-c} and \ref{fig:viz-d}, it is worth noting that some of the target objects are ignored by the ground truth labels, which are correctly located by the proposed method. This may lead to inferior performance due to inconsistent labelling.}
\label{fig:viz}
\end{figure*}

\subsection{Implementation}
The proposed method is built based on the widely used OpenPCDet \cite{openpcdet} code base. Particularly, we conduct experiments on two baseline detectors: Voxel-RCNN \cite{voxelrcnn} and PV-RCNN \cite{pvrcnn}. The corresponding frameworks are named as POP-RCNN-V and POP-RCNN-PV respectively. For the Waymo dataset, single frame input with only the first LiDAR return is used for all models.

\subsubsection{POP-RCNN-V}
POP-RCNN-V is built on Voxel-RCNN \cite{voxelrcnn} by replacing the Voxel RoI Pooling with POP-Pool, where the neighbouring voxel centres are used as the support points to represent the feature locations. Feature sources include the $2\times$, $4\times$ and $8\times$ voxel features as well as the BEV features. The DADCS scheme is also adopted to calibrate the confidence. Other configurations and components are kept the same.

\subsubsection{POP-RCNN-PV}
\label{sec:pv-implement}
POP-RCNN-PV is built on PV-RCNN \cite{pvrcnn} by replacing the RoI-grid Pooling with POP-Pool, where the sampled key points are still used to represent the feature locations.  While the Voxel Set Abstraction (VSA) in PV-RCNN requires more computational resources, we are not able to adopt the POP-Fuse module directly. Instead, the number of layers and the number of feature channels in the VSA module are reduced. In the original PV-RCNN \cite{pvrcnn}, all features aggregated with VSA to the key points via FPS are directly concatenated and processed with a fully connected layer for features output to the refinement stage. In contrast, POP-RCNN keeps the aggregated features separate according to the different feature sources by removing the concatenation operation and the fully connected layer. Instead, features from different scales are processed independently with POP-Pool, before being fed into the POP-Fuse module. This approach is feasible because the multi-scale nature of the features is not altered by the source-independent feature aggregation. The source-independent VSA is able to maintain the spatial stride of voxel-wise features, since only the accurate positional information is explicitly assigned to these voxel features. Feature sources only include raw points, the $4\times$ and $8\times$ voxel features and the BEV features.

For both frameworks, the POP-Pool module samples 4 levels of grid points to match the number of features. There are $(6,\;6,\;6)$, $(4,\;4,\;4)$, $(2,\;2,\;2)$ and $(2,\;2,\;2)$ grid points for each level in the order of the decreasing feature resolution. The POP-Fuse module also has 4 levels in spatial scales, 13 layers in semantic depths and 256 internal feature channels for each node. The each output node at the last layer has 60 output feature channels. This can be summarised as: $(L,\; D,\; C_i,\; C_o) = (4,\; 14,\; 256,\; 60)$. The shortcuts are built with $log_2n$ connections to reduce computational intensity. 

The model is trained with the ADAM optimiser on 4 RTX 2080 Ti GPUs. On the KITTI dataset, the model is trained with a batch size of 8 and a learning rate of 0.007 for 80 epochs. On Waymo Open dataset, the model is trained with a batch size of 8 and a learning rate of 0.007 for 40 epochs. Following OpenPCDet \cite{openpcdet}, the cosine annealing learning rate decay strategy is adopted with the data augmentation scheme unchanged, compared to those of the baseline models.

\subsection{Waymo Results}
We conduct experiments on Waymo Open Dataset to evaluate the effectiveness of POP-RCNN. The main results on Waymo Open dataset are shown in \tab\ref{tab:wod_obj}, where we compare the proposed method with the recent methods across two difficulty levels and three object categories. POP-RCNN shows improvements of 2.30\% and 0.84\% in terms of mAP scores on Vehicle LEVEL\_1 in comparison with Voxel-RCNN and PV-RCNN respectively. On the more difficult Vehicle LEVEL\_2, POP-RCNN outperforms its baselines by larger margins of 2.88\% and 1.12\% respectively. It is worth noting that POP-RCNN-V gives better predictions on the Cyclist category. Both POP-RCNN detector show competitive results compared to their competitors across all categories. In addition, POP-RCNN can provide more accurate detection on smaller targets. It should be noted that recent methods, like SST \cite{fan2022embracing}, M3DETR \cite{guan2022m3detr}, have achieved a significant increase in the detection accuracy, while leveraging multi-frame inputs. Multi-frame inputs induce extra computations and inference time. By comparing with the single-frame variant of the mentioned methods (denoted with 1f in \tab\ref{tab:wod_obj}), POP-RCNN shows advantageous results.

\begin{table}[t]
\centering
\caption{Results on KITTI \textit{val} set for car and cyclist classes, with average precision of 40 recall points (R40).}
\begin{tabular}{ccccccc}
\hline
\multicolumn{1}{c|}{\multirow{2}{*}{Methods}}       & \multicolumn{3}{c|}{Car}                   & \multicolumn{3}{c}{Cyclist} \\
\multicolumn{1}{c|}{}                               & Easy  & Mod.  & \multicolumn{1}{c|}{Hard}  & Easy    & Mod.    & Hard    \\ \hline
\multicolumn{7}{c}{RGB+LiDAR}    \\
\multicolumn{1}{c|}{CLOCs \cite{clocs}}             & 89.49 & 79.31 & \multicolumn{1}{c|}{77.36} & 87.57   & 67.92   & 63.67   \\
\multicolumn{1}{c|}{DVF \cite{dvf}}                 & 93.07 & 85.84 & \multicolumn{1}{c|}{83.13} & -       & -       & -       \\
\multicolumn{1}{c|}{CAT-Det \cite{catdet}}          & 90.12 & 81.46 & \multicolumn{1}{c|}{79.15} & 87.64   & 72.82   & 68.20   \\ 
\multicolumn{1}{c|}{SFD† \cite{sfd}}                & 95.47 & 88.56 & \multicolumn{1}{c|}{85.74} & -       & -       & -       \\\hline
\multicolumn{7}{c}{LiDAR}       \\
\multicolumn{1}{c|}{PointPillars \cite{pointpillar}} & 87.75 & 78.39 & \multicolumn{1}{c|}{75.18} & 81.57   & 62.94   & 58.98   \\
\multicolumn{1}{c|}{Part-A2 \cite{parta2}}          & 89.47 & 79.47 & \multicolumn{1}{c|}{78.54} & 73.07   & 88.31   & \textbf{70.20}   \\
\multicolumn{1}{c|}{SA-SSD \cite{sassd}}            & 92.23 & 84.30 & \multicolumn{1}{c|}{81.36} & -       & -       & -       \\
\multicolumn{1}{c|}{Voxel-RCNN \cite{voxelrcnn}}    & 92.38 & 85.29 & \multicolumn{1}{c|}{82.86} & -       & -       & -       \\
\multicolumn{1}{c|}{PV-RCNN \cite{pvrcnn}}          & 92.57 & 84.83 & \multicolumn{1}{c|}{82.69} & -       & -       & -       \\
\multicolumn{1}{c|}{VoTr-TSD \cite{votr}}           & 89.04 & 84.04 & \multicolumn{1}{c|}{79.14} & -       & -       & -       \\
\multicolumn{1}{c|}{Pyramid-PV \cite{pyramidrcnn}}  & 89.37 & 84.38 & \multicolumn{1}{c|}{78.84} & -       & -       & -       \\
\multicolumn{1}{c|}{PDV \cite{PDV}}                 & 92.56 & 85.29 & \multicolumn{1}{c|}{83.05} & 92.72   & 74.23   & 69.60   \\
\multicolumn{1}{c|}{SASA \cite{sasa}}               & 88.76 & 82.16 & \multicolumn{1}{c|}{77.16} & -       & -       & -       \\
\multicolumn{1}{c|}{SE-SSD \cite{sessd}}            & 92.13 & 82.64 & \multicolumn{1}{c|}{82.40} & -       & -       & -       \\
\multicolumn{1}{c|}{BtcDet \cite{btcdet}}           & \textbf{93.15} & \textbf{86.28} & \multicolumn{1}{c|}{\textbf{83.86}} & 91.45   & \textbf{74.70}   & 70.08   \\
\multicolumn{1}{c|}{SPG \cite{spg}}                 & 92.53 & 85.31 & \multicolumn{1}{c|}{82.82} & -       & -       & -       \\ \hline
\multicolumn{1}{c|}{Voxel-RCNN \cite{voxelrcnn}}    & 92.38 & 85.29 & \multicolumn{1}{c|}{82.86} & -       & -       & -       \\
\multicolumn{1}{c|}{\textbf{POP-RCNN-V}}           & 93.05 & 85.92 & \multicolumn{1}{c|}{83.53}    & -      & -      & -     \\\hline
\multicolumn{1}{c|}{PV-RCNN \cite{pvrcnn}}          & 92.57 & 84.83 & \multicolumn{1}{c|}{82.69} & 88.88       & 71.95       & 66.78       \\
\multicolumn{1}{c|}{\textbf{POP-RCNN-PV}}          & 92.71 & 85.49 & \multicolumn{1}{c|}{83.10}    & \textbf{94.56}      & 74.62      & 70.07      \\\hline
\end{tabular}
\label{tab:kitti-val}
\end{table}

To overcome the unfavoured distribution of points in the data representation, the main idea of the proposed model is to increase feature richness by promoting communications across spatial scales. While distance invariance of features is crucial in the detection of faraway targets, a distance-aware density confidence scoring scheme is adopted in the classification scores. Both mechanisms are beneficial to increasing the visibility of objects with sparse point clouds in the feature space. The features enriched with multi-scale connections are critical in improving the feature effectiveness, especially in a sparse data representation. The DADCS scheme provides additional distance-invariant information about the proposal bounding boxes and reweighs the classification scores. Our solution shows advantages over existing methods on the long distance detection. The detection results are also analysed by detection ranges, which are shown in \tabs\ref{tab:wod_veh_range}, \ref{tab:wod_ped_range} and \ref{tab:wod_cyc_range}. For vehicle detection, both POP-RCNN based methods provide competitive results in the close range, and show advantages in the long-range detection ($>30$m), demonstrating the improvement obtained by mitigating point sparsity with POP-Fuse and DADCS.

\subsection{KITTI Results}
The results on KITTI \textit{val} split are shown in \tab\ref{tab:kitti-val}. POP-RCNN-V follows its baseline on the training configuration, where only the Car category is considered. The POP-RCNN based variants improve the Car mAP scores by 0.63\% and 0.66\% in comparison with the baseline Voxel-RCNN and PV-RCNN methods respectively, for prediction tasks with moderate difficulties. In addition, POP-RCNN-PV boosts the mAP scores on the Cyclist category by 5.68\%, 2.67\% and 3.29\% for easy, moderate and difficult detection tasks, respectively. The improved results on smaller objects further prove the effectiveness of the proposed modules in processing sparse point clouds and solving the imbalance distribution of points inside the bounding boxes.

The POP-RCNN-PV framework is trained on the KITTI \textit{trainval} split for multi-class detection for submission to the KITTI test server. The results are shown in \tab\ref{tab:kitti-test}. The proposed POP-RCNN improves its baseline by 1.11\% and 3.25\% in the Car and Cyclist categories pertaining to moderate difficulty, while achieving competitive performances among multiple modality methods.

\begin{table}[t]
\centering
\caption{Results on KITTI \textit{test} set for car and cyclist classes, with average precision of 40 recall points (R40).}
\label{tab:kitti-test}
\begin{tabular}{ccccccc}
\hline
\multicolumn{1}{c|}{\multirow{2}{*}{Methods}}       & \multicolumn{3}{c|}{Car}                   & \multicolumn{3}{c}{Cyclist} \\
\multicolumn{1}{c|}{}                               & Easy  & Mod.  & \multicolumn{1}{c|}{Hard}  & Easy    & Mod.    & Hard    \\ \hline
\multicolumn{7}{c}{RGB+LiDAR}    \\
\multicolumn{1}{c|}{CLOCs \cite{clocs}}             & 88.94 & 80.67 & \multicolumn{1}{c|}{77.15} & -       & -       & -       \\
\multicolumn{1}{c|}{DVF \cite{dvf}}                 & 90.99 & 82.40 & \multicolumn{1}{c|}{77.37} & -       & -       & -       \\
\multicolumn{1}{c|}{CAT-Det \cite{catdet}}          & 89.87 & 81.32 & \multicolumn{1}{c|}{76.68} & 83.68   & 68.81   & 61.45   \\ 
\multicolumn{1}{c|}{SFD \cite{sfd}}                & 91.73 & 84.76 & \multicolumn{1}{c|}{85.74} & -       & -       & -       \\\hline
\multicolumn{7}{c}{LiDAR}        \\
\multicolumn{1}{c|}{PointPillars \cite{pointpillar}} & 82.58 & 74.31 & \multicolumn{1}{c|}{68.99} & 77.10   & 58.65   & 51.92   \\
\multicolumn{1}{c|}{Part-A2 \cite{parta2}}          & 87.81 & 78.49 & \multicolumn{1}{c|}{73.51} & 79.17   & 63.52   & 56.93   \\
\multicolumn{1}{c|}{Point-RCNN \cite{pointrcnn}}    & 86.96 & 75.64 & \multicolumn{1}{c|}{70.70} & 74.96   & 58.12   & 49.01   \\
\multicolumn{1}{c|}{3DSSD \cite{3dssd}}             & 88.36 & 79.57 & \multicolumn{1}{c|}{74.55} & 82.48   & 64.10   & 56.90   \\
\multicolumn{1}{c|}{SA-SSD \cite{sassd}}            & 88.75 & 79.79 & \multicolumn{1}{c|}{74.16} &         &         &         \\
\multicolumn{1}{c|}{Voxel-RCNN \cite{voxelrcnn}}    & 90.90 & 81.62 & \multicolumn{1}{c|}{77.06} & -       & -       & -       \\
\multicolumn{1}{c|}{PV-RCNN++ \cite{pvrcnn++}}      & 90.14 & 81.88 & \multicolumn{1}{c|}{77.15} &  82.22   & 67.44   & 60.04   \\
\multicolumn{1}{c|}{PDV \cite{PDV}}                 & 90.43 & 81.86 & \multicolumn{1}{c|}{77.36}  & 83.04   & 67.81   & 60.46   \\
\multicolumn{1}{c|}{VoxSet \cite{voxset}}           & 88.53 & 82.06 & \multicolumn{1}{c|}{77.46} & -       & -       & -       \\
\multicolumn{1}{c|}{VoTr-TSD \cite{votr}}           & 89.90 & 82.09 & \multicolumn{1}{c|}{79.14} & -       & -       & -       \\
\multicolumn{1}{c|}{Pyramid-PV \cite{pyramidrcnn}}  & 88.39 & 82.08 & \multicolumn{1}{c|}{77.49} & -       & -       & -       \\
\multicolumn{1}{c|}{SASA \cite{sasa}}               & 88.76 & 82.16 & \multicolumn{1}{c|}{77.16} & -       & -       & -       \\
\multicolumn{1}{c|}{SE-SSD \cite{sessd}}            & \textbf{91.94} & 82.54 & \multicolumn{1}{c|}{77.15} & -       & -            & -   \\
\multicolumn{1}{c|}{BtcDet \cite{btcdet}}           & 90.64 & \textbf{82.86} & \multicolumn{1}{c|}{78.09} & 82.81   & \textbf{68.68}   & \textbf{61.81}   \\

\multicolumn{1}{c|}{SPG \cite{spg}}                 & 90.50 & 82.13 & \multicolumn{1}{c|}{\textbf{78.90}} & -       & -       & -       \\ \hline
\multicolumn{1}{c|}{PV-RCNN \cite{pvrcnn}}          & 90.25 & 81.43 & \multicolumn{1}{c|}{76.82} & 78.60   & 63.71   & 57.65   \\ 
\multicolumn{1}{c|}{\textbf{POP-RCNN-PV}}                         & 91.02     & 82.54     & \multicolumn{1}{c|}{77.76 }    & \textbf{84.01}      & 66.96      & 60.23      \\ \hline
\end{tabular}
\end{table}

\subsection{Qualitative Results}
Visualisations for detection results with POP-RCNN-PV are presented in \fig\ref{fig:viz}. With richer context brought by POP-Pool and POP-Fuse, occluded targets can be located correctly. An example can be found on the left in \fig\ref{fig:viz-a} and \ref{fig:viz-c}.   \fig\ref{fig:viz-b} and \ref{fig:viz-c} also show the network capabilities of detecting distant small objects, owing to the adoption of both the POP-Fuse and DADCS modules. However, the given visualisations also reveal some false positives (FP). Most of the FPs are correct targets but not provided in the ground truth annotations. This inconsistency may inevitably harm the quantitative performance measured in mAP.

\begin{table*}[t]
\centering
\caption{Component comparison for POP-RCNN-PV on 10\% Waymo Open dataset for vehicle class by range.}
\begin{tabular}{c|c|c|c|ccc}
\hline
\multirow{2}{*}{Config.} & \multirow{2}{*}{Cls. Scoring} & \multirow{2}{*}{Pooling} & \multirow{2}{*}{Memory (MB)} & \multicolumn{3}{c}{Vehicle LEVEL\_2 mAP/mAPH}   \\
                 & &    &     & 0-30m         & 30-50m        & 50m-inf      \\ \hline

1            & density confidence prediction \cite{PDV}  & POP-Pool & 1718    & 90.74/90.31      & 67.42/67.01       & 39.87/39.03      \\

2   & DADCS & RoI grid pyramid \cite{pyramidrcnn} & 2134     & 90.40/90.12     & 68.23/67.31       & 40.10/39.39     \\ \hline

Ours          & DADCS & POP-Pool  & 1722   & 90.60/90.17   & 68.86/68.38   & 42.14/41.60 \\ \hline
\end{tabular}
\label{tab:wod_range_componets}
\end{table*}

\begin{table}[t]
\centering
\caption{Component analysis for POP-RCNN-PV on 10\% Waymo Open dataset. Pool, Fuse and DADCS denote POP-Pool, POP-Fuse and Density Confidence Scoring respectively.}
\begin{tabular}{c|ccc|ccc}
\hline
\multirow{2}{*}{Config.} & \multirow{2}{*}{Pool} & \multirow{2}{*}{Fuse} & \multirow{2}{*}{DADCS} &  \multicolumn{3}{c}{LEVEL\_2 mAPH} \\
  &            &            &            & Vehicle & Pedestrian & Cyclist \\ \hline
3 &            &            &            & 66.53   & 61.71     & 66.79   \\
4 & \checkmark &            &            & 66.33   & 61.95     & 66.83   \\
5 & \checkmark & \checkmark &            & 67.39   & 62.75     & 67.43   \\
Ours & \checkmark & \checkmark & \checkmark & \textbf{67.74}   & \textbf{63.62}     & \textbf{67.93}   \\ \hline
\end{tabular}
\label{tab:component}
\end{table}

\subsection{Ablation Study}
In this section, we compare the effectiveness of each component and variation of the network. All models for experiments are trained on the 10\% training set of Waymo Open Dataset, where training samples are uniformly selected from all training sequences in WOD. The comparisons are made on the more difficult LEVEL\_2 mAPH.

\subsubsection{Network Components}
Two comparisons based on the proposed POP-RCNN-PV are made by: (1) replacing the proposed DADCS module with the density confidence prediction in PDV \cite{PDV}; (2) replacing the proposed POP-Pool module with the RoI-grid Pyramid in Pyramid-RCNN \cite{pyramidrcnn}. The comparisons are summarised in \tab\ref{tab:wod_range_componets}. In the first comparison, it can be noticed that the DADCS greatly improves the network's ability to detect faraway targets by introducing distance-invariant density features. The advantage escalates as the detection distance increases. The second comparison illustrates the improvement by POP-Pool in terms of both memory consumption and detection accuracy. Since each level of the RoI-grid Pyramid gathers features from the same combined feature map, the duplicated features consume more computational memory. Because features from different spatial scales are not processed independently, each set of grid points contains features at all scales. This leads to sub-optimal performance from the subsequent POP-Fuse module, which relies on separate input of multi-scale features. It is also worth noting that a larger improvement is observed in long-distance detection. Hence the POP-Pool module is essential for achieving the full potential of the POP-Fuse module.

The effectiveness of each component is tested. We compare POP-RCNN-PV with the baseline PV-RCNN. The results are shown in \tab\ref{tab:component}. POP-Pool provides limited enhancement compared to the baseline PV-RCNN, as the grid point pyramid cannot efficiently increase the feature density by itself. However, with the integration of POP-Pool and POP-Fuse, there is an increase of 0.86\%, 1.04\% and 0.64\% on Vehicle, Pedestrian and Cyclist respectively. By considering the relation between point density and object distance with DADCS, we achieve another boost of 0.35\%, 0.87\% and 0.50\% on the three categories respectively. It is noticed that a larger improvement is seen for pedestrian detection. This is explained by the fact that pedestrian detection is more severely affected by the insufficient multi-scale information.

\begin{table}[t]
\centering
\caption{Effects of different configurations of POP-Fuse for POP-RCNN-PV on 10\% Waymo Open dataset. Depth denotes the number of layers, $C_i$ denotes the number of internal feature channels}
\begin{tabular}{c|c|c|ccc}
\hline
\multirow{2}{*}{ } & \multirow{2}{*}{Depth} & \multirow{2}{*}{$C_i$} & \multicolumn{3}{c}{LEVEL\_2 mAPH} \\
  &                  &             & Vehicle & Pedestrian & Cyclist \\ \hline
POP-Fuse-9 &    9    &     400     & 66.59   & 62.02     & 67.27   \\
POP-Fuse-11 &   11   &     320     & 67.22   & 62.35     & 67.81   \\
POP-Fuse-14 &   14   &     256     & \textbf{67.74}   & 63.62     & \textbf{67.93}   \\
POP-Fuse-16 &   16   &     216     & 67.46   & \textbf{63.68}     & 67.93   \\ 
POP-Fuse-20 &   20   &     160     & 67.54   & 63.13     & 67.33   \\\hline
\end{tabular}
\label{tab:fuse_config}
\end{table}

\begin{table}[t]
\centering
\caption{Comparison of Computational requirements by networks.}
\label{tab:runtime}
\begin{tabular}{c|c|c}
\hline
Methods      & Inference Time (ms)  & Memory Consumption (MB)  \\ \hline
Voxel-RCNN   &         72      &       1334 \\
POP-RCNN-V  &        142      &       1583 \\ \hline
PV-RCNN      &        191      &       2293 \\
POP-RCNN-PV &        206      &       1722 \\\hline
\end{tabular}
\end{table}

\subsubsection{Depth of POP-Fuse}
Experiments are conducted on the depth of POP-Fuse with reference to \cite{giraffedet}. As the computational resource is limited, we keep the memory consumption of the POP-Fuse approximately constant. This is achieved by dynamically adjusting the number of internal feature channels according to the depth. The results are shown in \tab\ref{tab:fuse_config}. We adopt the configuration of 14 layers and 256 channels for achieving the best results.

\subsubsection{Computational Headroom}
By measuring the inference time and memory consumption of POP-RCNN and its baselines, a comparison of the network efficiency is shown in \tab\ref{tab:runtime}. The inference time and memory consumption are analysed by the average values of processing one sample in the \textit{val} set in KITTI dataset. POP-RCNN-V requires more computation headroom while achieving higher accuracies. POP-RCNN-PV requires less memory with the additional modules proposed in the paper, due to the simplified VSA module mentioned in Section \ref{sec:pv-implement}. POP-RCNN-PV provides superior performance with a only 7.9\% longer inference time. Both quantities are measured with a single GTX 1080 GPU. A reduction in running time is expected to realise real-time inference with high-end devices.

\section{Conclusion}
A two-stage detection framework, POP-RCNN, with a pyramid feature fusion scheme is presented to mitigate point cloud sparsity. The proposed framework can be adopted by a wide range of existing methods to increase the efficiency of fusing multi-scale features. The features with rich geometric and contextual information provide important clues to locate small and distant targets. Competitive performances are achieved on Waymo Open Dataset and KITTI dataset with simplified feature aggregation and limited computational resources. In the large and complex Waymo Open Dataset, both variations of POP-RCNN outperform their baselines. It is notable that the proposed model shows more significance as the distance increases. While the proposed model achieves competitive performance in close-range detection, more substantial improvement is shown by the increase of 1.25\% and 2.02\% in mAP on LEVEL\_1 Vehicle in the ranges of 30-50m and 50m-inf respectively. For the popular KITTI test benchmark, the variant method, \ie POP-RCNN-PV exceeds its baseline, \ie PV-RCNN, on the moderate Car and Cyclist categories by 1.11\% and 3.25\% respectively. By comparing the effects of each components in the model, it can be observed that the proposed network also consumes less computational resources in terms of memory usage by using a simplified feature aggregator in the backbone network. In the future, we would focus on the scalability of the framework, and aim to achieve better results with a larger network capacity.


\bibliography{bibli}
\bibliographystyle{IEEEtran}








\end{document}